\documentclass[10pt,twocolumn,letterpaper]{article}

\usepackage[article]{cvpr}

\usepackage{graphicx}
\usepackage{amsmath}
\usepackage{amssymb}
\usepackage{booktabs}
\usepackage{ctable}
\usepackage{multirow}
\usepackage{array}
\usepackage{colortbl}
\usepackage{booktabs}
\usepackage[utf8]{inputenc}
\usepackage{newunicodechar}
\newunicodechar{∆}{\Delta}
\usepackage{tabularx}
\usepackage{array}
\usepackage{makecell}
\usepackage{microtype}

\definecolor{cvprblue}{rgb}{0.21,0.49,0.74}
\usepackage[pagebackref,breaklinks,colorlinks,citecolor=cvprblue]{hyperref}

\def\paperID{24}
\def\confName{CVPR}
\def\confYear{2026}

\begin{document}
	
	\title{MobileAgeNet: Lightweight Facial Age Estimation for Mobile Deployment}
	
	\author{Arun Kumar{\thanks{Corresponding author: arunkumar03.de@gmail.com}}, \space\space\space Aswathy Baiju,\space\space\space Radu Timofte,\space\space\space Dmitry Ignatov$^{\dagger}$\\
		\small{Computer Vision Lab, CAIDAS \& IFI, University of W\"urzburg, Germany $\cdot$ $^{\dagger}$dmytro.ignatov@uni-wuerzburg.de}}
	\maketitle
	
\begin{abstract}
Mobile deployment of facial age estimation requires models that balance 
predictive accuracy with low latency and compact size. In this work, we 
present MobileAgeNet, a lightweight age-regression framework that achieves 
MAE of 4.65 years on UTKFace held-out test set while maintaining efficient 
on-device inference with an average latency of 14.4 ms measured using the 
AI Benchmark application. The model is built on a pretrained 
MobileNetV3-Large backbone combined with a compact regression head, 
enabling real-time prediction on mobile devices. The training and evaluation 
pipeline is integrated into the NN LEMUR Dataset 
{\small(\url{https://github.com/ABrain-One/NN-Dataset})} framework, 
supporting reproducible experimentation, structured hyperparameter 
optimization, and consistent evaluation. We employ bounded age regression 
together with a two-stage fine-tuning strategy to improve training stability 
and generalization. Experimental results show that MobileAgeNet achieves 
competitive accuracy at 3.23M parameters, and that the deployment 
pipeline---from PyTorch training through ONNX export to TensorFlow Lite 
conversion---preserves predictive behavior without measurable degradation 
under practical on-device conditions. Overall, this work provides a 
practical, deployment-ready baseline for mobile-oriented facial age 
estimation.
\end{abstract}
\section{Introduction}
\label{sec:intro}

Facial age estimation aims to predict a person’s age from facial imagery and remains a challenging problem in computer vision and biometrics \cite{Fu2010AgeSurvey,Angulu2018AgeSurvey}. The difficulty arises from the highly individual nature of facial aging, as well as variations in pose, illumination, expression, occlusion, and image quality. In addition, ambiguity between neighboring ages further complicates prediction \cite{Angulu2018AgeSurvey,Wen2020AVDL}. Recent work has also shown that evaluation protocols, including dataset splits and preprocessing choices, can significantly affect reported performance, highlighting the importance of reproducible evaluation \cite{Paplham2024Benchmark}.

For practical applications, particularly on mobile and edge devices, efficiency is as important as predictive accuracy. Large convolutional models can achieve strong performance, but their computational cost, memory footprint, and inference latency often limit their usability in resource-constrained environments. MobileNetV3 was designed to address this trade-off by providing efficient visual representations tailored for mobile deployment \cite{Howard2019MobileNetV3}. Prior work has likewise demonstrated that compact age-estimation models can remain competitive without relying on heavy architectures \cite{Zhang2019C3AE,Savchenko2019MultiOutputCNN}.

Motivated by these considerations, we propose a lightweight facial age-estimation pipeline based on a pretrained MobileNetV3-Large backbone and a compact regression head. The model performs continuous age prediction using bounded regression and a staged fine-tuning strategy. This design enables stable optimization while preserving efficiency, and leverages pretrained visual representations without introducing additional architectural complexity.

A key aspect of this work is reproducibility. The proposed pipeline is implemented within the NN LEMUR Dataset framework \cite{Goodarzi2025LEMUR}, which provides a unified environment for dataset integration, model training, hyperparameter optimization, and evaluation. This setup enables consistent experimental protocols, including validation-based checkpoint selection and held-out testing, thereby reducing variability in reported results.

We evaluate the proposed approach on UTKFace, a widely used dataset containing over 20{,}000 images spanning ages from 0 to 116 years \cite{UTKFaceDataset}. Our goal is to study how a lightweight pretrained backbone performs under realistic visual variability while maintaining deployment efficiency.

In summary, this work makes three contributions. First, we present a lightweight age-estimation model suitable for mobile deployment based on MobileNetV3-Large. Second, we provide a reproducible training and evaluation pipeline integrated into NN LEMUR Dataset framework. Third, we demonstrate that a compact pretrained backbone can achieve competitive performance on UTKFace while maintaining low computational cost and real-time inference capability.

\section{Related Work}
\label{sec:Related}

\subsection{Facial Age Estimation}
\label{subsec:related_age_estimation}

Facial age estimation has been studied for many years and remains a challenging problem due to the variability of age-related facial cues across individuals. These cues are strongly influenced by pose, illumination, occlusion, expression, image quality, and annotation ambiguity \cite{Fu2010AgeSurvey,Angulu2018AgeSurvey}. Public benchmarks have played an important role in advancing the field. Datasets such as Adience, AgeDB, and UTKFace enable evaluation under diverse conditions, with UTKFace being widely used due to its broad age range and significant visual variability \cite{Eidinger2014Adience,Moschoglou2017AgeDB,UTKFaceDataset}.

A major line of work focuses on how age should be modeled during learning. Since age is inherently ordered, several approaches formulate age estimation as an ordinal prediction problem rather than standard classification, including ordinal regression and ranking-based CNN methods \cite{Niu2016ORCNN,Chen2017RankingCNN}. Other studies explicitly model age uncertainty by assigning a distribution over neighboring ages, as explored in label distribution learning and mean-variance based approaches \cite{He2017DDLDL,Pan2018MeanVariance,Wen2020AVDL}.

In addition, regression-based formulations aim to capture the continuous nature of age while maintaining training stability \cite{Shen2018DeepRegressionForests,Li2019BridgeNet}. More recent methods incorporate auxiliary cues and localized facial modeling. Demographic-assisted cascaded structures refine predictions using additional information \cite{Wan2018CascadedStructure}, while attention-based and region-focused architectures emphasize age-sensitive facial regions \cite{Zhang2020AttentionLSTM,Wang2022DynamicPatchFusion}. Transformer-based designs have also been explored to encode richer representations and model refined age transitions \cite{Chen2023DAA}. Although these approaches improve predictive performance, they often introduce increased architectural complexity.

\subsection{Lightweight Architectures and Reproducible Evaluation}
\label{subsec:related_lightweight_evaluation}

Alongside predictive accuracy, computational efficiency is critical for practical age-estimation systems, particularly for deployment on mobile and edge devices. Large convolutional backbones can achieve strong results, but their computational cost, memory footprint, and inference latency often limit their usability in resource-constrained environments. Large-scale visual pretraining remains a common foundation for transfer learning in facial analysis \cite{Deng2009ImageNet}, making mobile-oriented architectures attractive candidates for efficient age estimation. MobileNetV3 is especially relevant in this context, as it was designed to optimize the accuracy--latency trade-off for mobile vision applications and has also been studied as an effective architecture for lightweight image classification \cite{Howard2019MobileNetV3,Qian2021MobileNetV3Classification}.

In the age-estimation literature, compact architectures have received increasing attention. C3AE demonstrated that strong performance can be achieved with a deliberately lightweight design \cite{Zhang2019C3AE}, while efficient multi-output CNNs have been explored for joint facial attribute prediction, including age estimation \cite{Savchenko2019MultiOutputCNN}. These studies motivate the use of compact pretrained backbones as a practical alternative to heavier age-estimation pipelines.

Recent work has also emphasized that reported performance can depend strongly on the evaluation protocol. Dataset properties, preprocessing choices, split design, and checkpoint-selection strategies can all influence results and make fair comparison difficult \cite{Paplham2024Benchmark}. This is particularly relevant when comparing methods across datasets such as Adience, AgeDB, and UTKFace, which differ in scale, collection process, and visual variability \cite{Eidinger2014Adience,Moschoglou2017AgeDB,UTKFaceDataset}. As a result, reproducible training and evaluation pipelines are increasingly important in age-estimation research.

Our work is most closely related to approaches that aim to balance efficiency and predictive performance. Unlike methods that rely on heavy attention mechanisms, complex label-distribution heads, or specialized transformer-based modules \cite{Pan2018MeanVariance,Wen2020AVDL,Zhang2020AttentionLSTM,Wang2022DynamicPatchFusion,Chen2023DAA}, we focus on a lightweight regression pipeline built on a pretrained MobileNetV3-Large backbone with a compact prediction head. In addition, we emphasize reproducibility through structured dataset preparation, validation-based checkpoint selection, held-out evaluation, and integration into the NN LEMUR Dataset framework\cite{Howard2019MobileNetV3,Qian2021MobileNetV3Classification,Zhang2019C3AE,Paplham2024Benchmark,Goodarzi2025LEMUR}. This positions the proposed method as a practical and deployment-oriented baseline for facial age estimation under a transparent and reproducible evaluation setting.

Overall, this section highlights the evolution of age-estimation methods from complex modeling strategies toward efficient and reproducible designs, motivating the lightweight and deployment-oriented approach adopted in this work.

\section{Methodology}
\label{sec:Methodology}

This section presents the data preparation procedure, the proposed MobileAgeNet architecture, the staged training strategy, and the hyperparameter optimization process implemented within the NN LEMUR Dataset framework \cite{Goodarzi2025LEMUR}.

\subsection{Dataset Preparation}
\label{subsec:data_preparation}

Facial age estimation was formulated as a supervised regression task in which each facial image was associated with a continuous age label. Before training, the dataset was curated to remove unreliable samples. Entries with missing images or missing age annotations were discarded, and only samples with valid numeric ages in the range 0--116 years were retained. This filtering step helped reduce the influence of corrupted or implausible labels on subsequent model development.

To ensure a consistent visual representation, all images were converted to RGB format when necessary. The preprocessing pipeline then applied channel-wise normalization with mean values of $(0.485, 0.456, 0.406)$ and standard deviations of $(0.229, 0.224, 0.225)$, followed by conversion to tensor form for training in PyTorch. Age labels were stored as floating-point targets, in accordance with the regression formulation of the task.

To preserve the age distribution across experimental subsets, the dataset was partitioned using stratified sampling based on 5-year age bins. Each sample was assigned to a 5-year age bin, and the data were split within each bin into training, validation, and test subsets with approximate proportions of 70\%, 10\%, and 20\%, respectively. This strategy reduced the likelihood that particular age groups would be overrepresented or underrepresented in any single subset and provided a more reliable basis for model selection and final performance assessment. A fixed random seed was used to make the partitioning process reproducible.

The validation subset was used for model selection, while a separate held-out test subset was reserved exclusively for final evaluation. As a result, model selection and checkpoint retention were performed without access to the final test partition. In addition, imbalance-aware sample weights were computed for the training subset using the inverse square root of the corresponding age-bin frequency. These weights were stored as dataset metadata for potential future use in weighted sampling or loss reweighting.

\subsection{Proposed Model Architecture}
\label{subsec:model_architecture}

\begin{figure*}[t]
    \centering
    \includegraphics[width=0.72\textwidth, trim=8 8 8 8, clip]{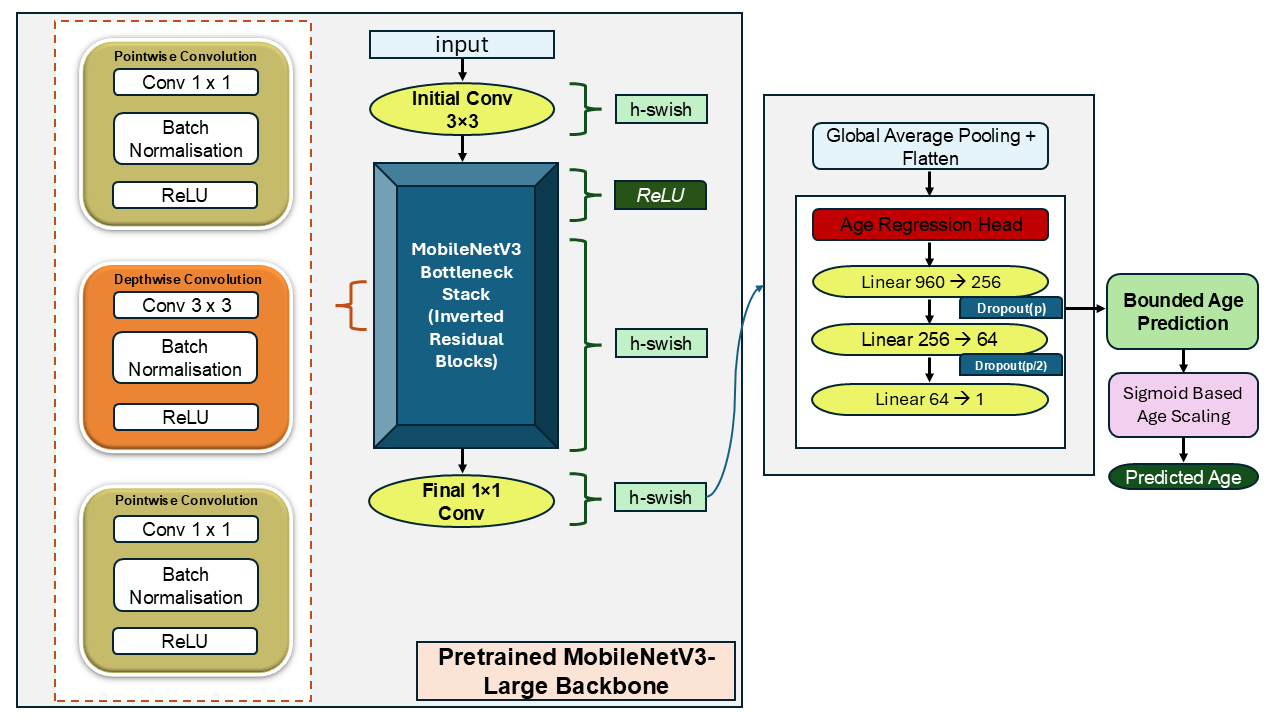}
    \caption{Architecture of the proposed MobileAgeNet. A pretrained MobileNetV3-Large backbone~\cite{Qian2021MobileNetV3Classification} is used for feature extraction, followed by global average pooling and flattening. The resulting feature vector is passed to a lightweight age regression head with fully connected layers ($960 \rightarrow 256 \rightarrow 64 \rightarrow 1$), and the final scalar output is mapped to a bounded age range using sigmoid-based age scaling.}
    \label{fig:mobilenetv3_backbone}
    \vspace{-3mm}
\end{figure*}

The proposed model, referred to as \textit{MobileAgeNet}, is a lightweight facial age regressor built on top of a pretrained MobileNetV3-Large backbone~\cite{Howard2019MobileNetV3}. The overall architecture is illustrated in Fig.~\ref{fig:mobilenetv3_backbone}. The model follows a compact end-to-end design consisting of a convolutional feature extractor, a global pooling stage, a lightweight regression head, and a bounded output mapping.

The convolutional backbone inherits the MobileNetV3-Large architecture, including the initial convolution, a stack of inverted residual bottleneck blocks, and a final pointwise convolution. These components provide efficient feature extraction using depthwise separable convolutions and nonlinear activations such as ReLU and h-swish. The pretrained backbone enables the model to leverage expressive visual representations learned from large-scale ImageNet pretraining~\cite{Deng2009ImageNet} while maintaining low computational cost.

Following the backbone, global average pooling is applied to aggregate spatial features into a compact representation, which is then flattened and passed to a lightweight regression head. The regression head consists of two fully connected layers with dimensions $960 \rightarrow 256$ and $256 \rightarrow 64$, each followed by Hardswish (h-swish) activation and dropout regularization. A final linear layer maps the learned representation to a single scalar output corresponding to age. The linear layers were initialized using truncated normal initialization, with biases initialized to zero.

To ensure realistic predictions, the model employs a bounded-output formulation. The scalar output of the regression head is passed through a sigmoid mapping and scaled to a predefined age range. In the present implementation, the overall supported range is $[0,116]$ years, while the effective training range is restricted to $[1,95]$ years. This bounded formulation reduces the likelihood of extreme predictions and improves training stability by discouraging implausible outputs.

\subsection{Training Strategy}
\label{subsec:training_strategy}

Training was carried out using a two-stage transfer-learning strategy. In the first stage, the convolutional backbone was frozen and only the regression head was optimized. This warm-up stage allowed the newly introduced prediction layers to adapt to the age-estimation task before the pretrained backbone was fine-tuned. During this stage, batch-normalization layers in the backbone were kept in evaluation mode in order to improve transfer-learning stability.

After the warm-up period, the backbone was unfrozen and the model entered a fine-tuning stage in which the backbone and regression head were optimized jointly. Different learning rates were assigned to the two parameter groups: the regression head used the base learning rate, while the backbone used a reduced learning rate scaled by a factor of $0.10$. This design allowed the pretrained feature extractor to adapt gradually while preserving the benefits of initialization from large-scale pretraining.

Age prediction was optimized as a regression problem using the Smooth L1 loss with $\beta = 1.0$. During training, target ages were clamped to the effective bounded training range so that the supervision remained consistent with the model output constraint. Optimization was performed with AdamW, and cosine annealing was used to schedule the learning rate during both the warm-up and fine-tuning stages. To further stabilize optimization, gradient clipping with a maximum norm of 2.0 was applied during backpropagation.

\subsection{Hyperparameter Optimization}
\label{subsec:hyperparameter_optimization}

Hyperparameter selection was performed in two phases. In the first phase, an automated search procedure based on Optuna was used to identify suitable training settings for MobileAgeNet. The search space included the learning rate, dropout probability, batch size, and image transformation strategy, while several model-specific parameters were fixed in advance, including the number of frozen epochs, the backbone learning-rate multiplier, and the bounded target range.

The learning rate was searched on a logarithmic scale between $5 \times 10^{-4}$ and $2 \times 10^{-3}$, dropout was searched in the interval $[0.10, 0.30]$, and the batch size was selected from powers of two corresponding to 64 and 128 samples per batch. The transformation search was intentionally restricted to face-suitable augmentation pipelines, prioritizing normalization, resizing, horizontal flipping, and moderate appearance perturbations over more aggressive transformations that could distort age-relevant facial cues.

In the second phase, the best-performing hyperparameter configuration identified during the search was fixed and used for final training. This locked-parameter training stage was executed as a single final run with checkpoint saving enabled, thereby separating exploratory search from the final reported model.

\section{Experiments}
\label{sec:experiments}

This section describes the experimental protocol used to evaluate the proposed MobileAgeNet framework on UTKFace. The experiments are designed around two practical objectives: first, to assess whether a lightweight pretrained backbone can support accurate facial age estimation, and second, to ensure that model selection and final reporting follow a reproducible procedure \cite{Howard2019MobileNetV3,Zhang2019C3AE,Paplham2024Benchmark,Goodarzi2025LEMUR}.

\subsection{Experimental Setup}
\label{subsec:experimental_setup}

All experiments were conducted within the NN LEMUR Dataset framework using UTKFace as the target dataset and MobileAgeNet as the prediction model. The experimental pipeline was organized in two phases. The first phase was used for hyperparameter search, while the second phase was reserved for final model training with a fixed configuration. This separation reduces the risk of tuning and reporting on the same run and ensures that the final result corresponds to a single locked setting rather than exploratory experimentation \cite{Paplham2024Benchmark}.

Following the search space defined in Sect.~\ref{subsec:hyperparameter_optimization}, we employed Optuna to systematically explore key hyperparameters, including the learning rate, dropout probability, batch size, and the data transformation pipeline. The augmentation search space was deliberately constrained to a limited set of face-appropriate preprocessing strategies in order to avoid transformations that could distort age-relevant visual cues. The optimization procedure was conducted for up to 40 trials, with each trial trained for a maximum of 60 epochs.

The best configuration identified during the search was then fixed for the final experiment. In the current implementation, the final run used a learning rate of 0.0014162, a batch size of 64, a dropout value of 0.1807, a backbone learning-rate multiplier of 0.10, five frozen warm-up epochs, bounded output enabled, and the \texttt{Resize\_ColorJit\_Flip\_Blur} transformation pipeline. Final training was carried out for up to 100 epochs. This configuration defines the experimental setting reported in this work.

\subsection{Training and Testing}
\label{subsec:training_testing}

Model training followed a staged transfer-learning protocol in which the final locked configuration was optimized on the training subset and monitored on the validation subset. All major design choices were fixed prior to this stage, making the final run confirmatory rather than exploratory.

Validation performance was evaluated after each epoch and used for model selection. The best checkpoint observed during training was retained and restored at the end of the run. Final testing was performed once on the held-out test partition, which remained separate from both parameter updates and checkpoint selection. This design enforces a clear separation between model development and final generalization assessment \cite{Paplham2024Benchmark}.

The training framework also recorded additional diagnostic information. Per-epoch logs included training and validation loss, mean absolute error when available, current learning rate, gradient norm, and throughput in samples per second. TensorBoard visualizations were generated to support qualitative analysis, including predicted-versus-true scatter plots, prediction histograms, and sample-image summaries. These diagnostics are particularly useful for lightweight models, where optimization stability and computational behavior are as important as final accuracy \cite{Howard2019MobileNetV3,Zhang2019C3AE}.

\subsection{Evaluation Metrics}
\label{subsec:evaluation_metrics}

The primary evaluation metric is mean absolute error (MAE), reported in years. MAE is widely used in facial age estimation because it measures the average absolute deviation between predicted and ground-truth ages in a directly interpretable form \cite{Rothe2018DEX,Pan2018MeanVariance,Wen2020AVDL,Paplham2024Benchmark}. It is also well aligned with the regression-based formulation of MobileAgeNet, which predicts a single continuous age value.

In addition to MAE, the framework tracks loss values throughout training and validation to monitor convergence and identify the best checkpoint. Auxiliary statistics such as gradient norm, throughput, and resource usage are also recorded. While these are not primary evaluation metrics, they provide useful context for understanding the behavior of a lightweight model intended for efficient deployment.

Given the focus on both predictive performance and deployment efficiency, MAE is used as the main reported metric, while auxiliary statistics support analysis of training dynamics and computational cost. This reporting strategy is consistent with recent work advocating more transparent evaluation practices in age-estimation research \cite{Paplham2024Benchmark}.

\section{Ablation Study}
\label{sec:ablation}

We report two focused ablations that had the strongest impact on model performance. All ablation results correspond to intermediate configurations explored prior to the final hyperparameter tuning stage and therefore do not match the final reported model performance. Table~\ref{tab:ablation} summarizes the results.

\begin{table}[t]
\centering
\small
\caption{Ablation results on UTKFace. Lower MAE is better.}
\begin{tabular}{lc}
\toprule
Setting & Test MAE ($\downarrow$) \\
\midrule
\texttt{norm\_256} & 5.568 \\
\texttt{norm\_256\_flip} & \textbf{4.701} \\
\bottomrule
\end{tabular}
\label{tab:ablation}
\end{table}

The first ablation compares the earlier training pipeline with the final pipeline. The earlier version used a simpler regression head, a longer frozen stage with \texttt{freeze\_epochs} = 10, a smaller backbone learning-rate multiplier with \texttt{backbone\_lr\_mult} = 0.05, and a large positive output-layer bias before sigmoid-based age bounding. In the revised pipeline, this design was replaced by a more stable bounded-regression formulation, corrected output initialization, a shorter warm-up stage, and stronger backbone adaptation during fine-tuning. This change produced a substantial improvement, with the revised pipeline achieving a held-out MAE of 4.701.

The second ablation examines the effect of augmentation by comparing \texttt{norm\_256} and \texttt{norm\_256\_flip} within the revised pipeline. A representative \texttt{norm\_256} run reached an MAE of 5.568, whereas the \texttt{norm\_256\_flip} configuration achieved a held-out MAE of 4.701. This suggests that horizontal flipping improves robustness while preserving age-relevant facial structure.

Taken together, the ablation results indicate that the main gains come from stabilizing the regression and fine-tuning pipeline, followed by selecting a simple but effective augmentation strategy. Importantly, these improvements are achieved without replacing the lightweight MobileNetV3 backbone.
\section{Results and Discussion}
\label{sec:results}

We report results on the UTKFace dataset using the two-phase pipeline described earlier. Phase~1 was used for Optuna-based hyperparameter search, and Phase~2 used the best configuration from Phase~1 for final locked training and held-out evaluation. Since the task is formulated as age regression, mean absolute error (MAE) in years is treated as the primary performance metric.

\begin{figure*}[t]
    \centering
    \includegraphics[width=0.84\textwidth,height=0.28\textheight,keepaspectratio]{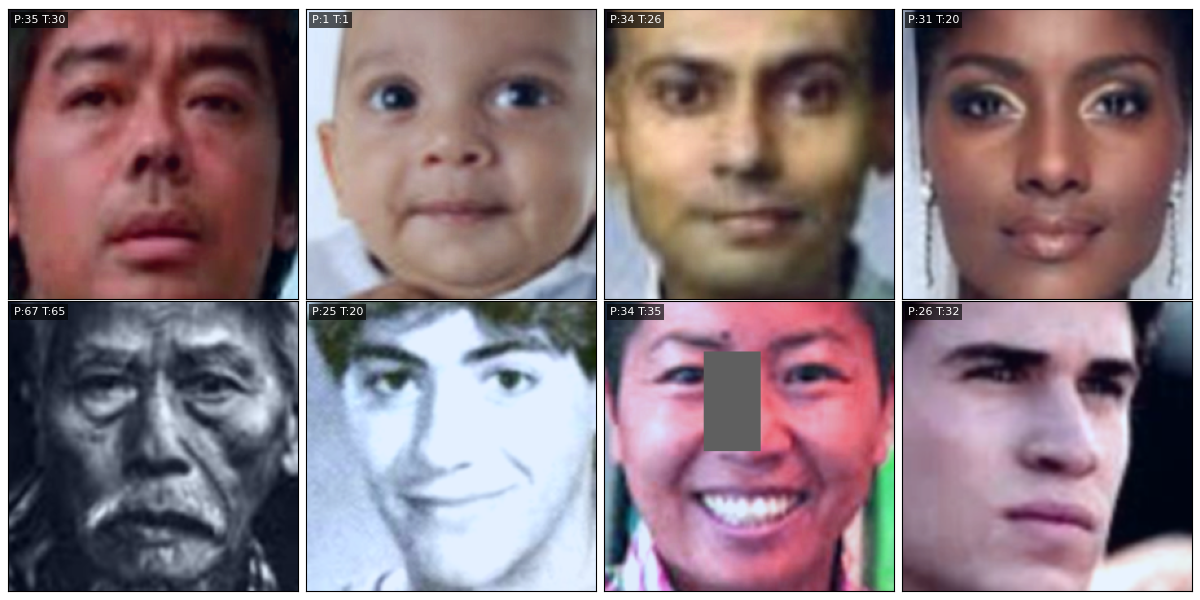}
    \caption{Qualitative results from the best checkpoint (epoch 95), showing predicted age (P) and ground-truth age (T).}
    \label{fig:qualitative_results}
    \vspace{-2mm}
\end{figure*}

\subsection{Phase-1 Hyperparameter Search}
\label{subsec:phase1_results}

Phase~1 was used to identify a suitable training configuration for MobileAgeNet on UTKFace before running the final locked experiment. Hyperparameter search was carried out with Optuna over 40 trials. Each trial was trained for up to 60 epochs and evaluated using the same validation protocol and held-out test split defined for the dataset. The search space covered the learning rate, dropout probability, batch size, and image transformation pipeline, while the remaining model-specific settings were kept fixed. This design allowed the search phase to focus on the most influential optimization and regularization parameters without changing the underlying model formulation.

Table~\ref{tab:phase1_results} reports the best-performing trial from this search. Trial 34 achieved an MAE of 4.75 years and was selected as the basis for Phase~2, where its hyperparameter configuration was fixed and used for the final training run. The Phase~1 results therefore serve both as the basis for selecting the final configuration and as evidence that the proposed MobileAgeNet pipeline performs well already during the search stage.

\begin{table}[t]
\centering
\small
\caption{Phase-1 hyperparameter search result on UTKFace. Optuna was run for 40 trials, and each trial was trained for up to 60 epochs. The table reports the best-performing trial selected for the final Phase-2 run.}
\begin{tabular}{lcc}
\toprule
Trial & Model & MAE ($\downarrow$) \\
\midrule
34 & MobileAgeNet & 4.75 \\
\bottomrule
\end{tabular}
\label{tab:phase1_results}
\end{table}

\subsection{Phase-2 Final Training}
\label{subsec:phase2_results}

In Phase~2, the best hyperparameter configuration identified during the Optuna search was fixed and used for the final training run. This stage was intended to provide a single reproducible model instance using the selected settings, rather than to continue exploratory tuning. The final configuration used a learning rate of 0.0014162, batch size 64, dropout 0.18074, and a backbone learning-rate multiplier of 0.10. As in the search stage, the model was trained under the same validation protocol, and the best checkpoint observed during training was retained as the final model for evaluation.

Table~\ref{tab:phase2_results} summarizes the final Phase-2 outcome. The final run was carried out for 100 epochs, and the best checkpoint was obtained at epoch 95. This checkpoint achieved a held-out MAE of 4.65 years, improving upon the best result observed during Phase~1. The close agreement between the Phase~1 trial and the Phase~2 checkpoint indicates that the selected hyperparameter configuration transfers reliably from the search stage to the final locked run.

The reported Phase~2 result corresponds to the best checkpoint selected during training rather than to the final epoch. This reflects the model instance retained by the training pipeline for final testing and supports the suitability of MobileAgeNet as a lightweight age-regression model with competitive performance and a compact deployment-oriented design.

\begin{table}[t]
\centering
\small
\setlength{\tabcolsep}{4pt}
\caption{Phase-2 final training and testing summary on UTKFace. The table reports the locked hyperparameter configuration selected from Phase~1 and the best checkpoint obtained during the final 100-epoch training run.}
\begin{tabularx}{\columnwidth}{@{}l>{\raggedright\arraybackslash}X@{}}
\toprule
Metric & Value \\
\midrule
Model & MobileAgeNet \\
Learning rate & 0.0014162 \\
Batch size & 64 \\
Dropout & 0.18074 \\
Backbone LR multiplier & 0.10 \\
Epochs & 100 \\
Best checkpoint epoch & 95 \\
Held-out MAE ($\downarrow$) & 4.65 years \\
\bottomrule
\end{tabularx}
\label{tab:phase2_results}
\end{table}

\subsection{Contextual Comparison with Lightweight and Benchmark Baselines}
\label{subsec:comparison_baselines}

Table~\ref{tab:comparison_baselines} places the proposed MobileAgeNet result in the context of existing UTKFace results. Because UTKFace is evaluated under different data splits, age ranges, and preprocessing protocols across the literature, these comparisons should be interpreted with caution. In particular, recent benchmark analysis has shown that protocol differences can substantially affect reported age-estimation performance \cite{Paplham2024Benchmark}. For this reason, we separate full-range lightweight references, restricted-split lightweight references, and standardized benchmark references rather than treating all rows as directly comparable.

Within the class of lightweight models evaluated on UTKFace, the proposed MobileAgeNet achieves a held-out MAE of 4.65 years. This is numerically lower than the MobileNet-based results reported by Savchenko in the UTKFace setting of that paper, which include MAE values of 5.44 and 5.74 \cite{Savchenko2021Lightweight}. More recent distilled MobileNetV3 student models also provide relevant efficiency-oriented references: distilled MobileNetV3-Large and MobileNetV3-Small achieved MAE values of 5.0033 and 5.0693, respectively \cite{Kim2024KD}. However, those results were obtained on a restricted UTKFace subset containing ages 21--60 only, with 13{,}144 training images and 3{,}287 test images, and therefore should not be interpreted as directly comparable to the full-range setting used in this work \cite{Kim2024KD}.

For broader context, Table~\ref{tab:comparison_baselines} also includes two stronger reference points from the unified CVPR 2024 benchmark. Under that benchmark protocol, EfficientNet-B4 achieved 4.23 MAE and FaRL+MLP achieved 3.87 MAE on UTKFace \cite{Paplham2024Benchmark}. These models are useful as benchmark reference points, but they are not direct mobile-oriented baselines. Overall, the comparison indicates that MobileAgeNet achieves competitive performance within the class of lightweight age-estimation models while maintaining a compact and deployment-oriented design.

The final model contains approximately 3.23M parameters and requires 0.233 GFLOPs per inference, making it suitable for deployment on resource-constrained devices. This highlights the favorable balance between predictive performance and computational efficiency achieved by the proposed approach in comparison to both lightweight and larger benchmark models.

\begin{table}[t]
\centering
\small
\setlength{\tabcolsep}{3.2pt}
\caption{Contextual comparison on UTKFace. Lower MAE is better. Results are grouped by evaluation setting due to differences in splits and preprocessing protocols.}
\begin{tabularx}{\columnwidth}{@{}>{\raggedright\arraybackslash}X>{\raggedright\arraybackslash}Xc>{\raggedright\arraybackslash}X@{}}
\toprule
Method & Setting & MAE ($\downarrow$) & Params / GFLOPs \\
\midrule
\multicolumn{4}{@{}l}{\textit{Full-range UTKFace (lightweight)}} \\
MobileAgeNet (ours) & Full range, stratified split & 4.65 & 3.23M / 0.233G \\
Savchenko MobileNet~\cite{Savchenko2021Lightweight} & Full range, fine-tuned & 5.44 & not reported \\
Savchenko MobileNet~\cite{Savchenko2021Lightweight} & Full range, VGGFace2 pretrain & 5.74 & not reported \\
\midrule
\multicolumn{4}{@{}l}{\textit{Restricted UTKFace subset (lightweight)}} \\
Distilled MobileNetV3-Large~\cite{Kim2024KD} & Ages 21--60, 13k/3k split & 5.0033 & 3.89M / 0.220G \\
Distilled MobileNetV3-Small~\cite{Kim2024KD} & Ages 21--60, 13k/3k split & 5.0693 & 1.48M / 0.057G \\
\midrule
\multicolumn{4}{@{}l}{\textit{Unified benchmark}} \\
EfficientNet-B4~\cite{Paplham2024Benchmark} & Standardized CVPR'24 protocol & 4.23 & not reported \\
FaRL + MLP~\cite{Paplham2024Benchmark} & Standardized CVPR'24 protocol & 3.87 & not reported \\
\bottomrule
\end{tabularx}
\label{tab:comparison_baselines}
\end{table}

\subsection{Deployment Consistency from ONNX to TFLite}
\label{subsec:deployment_consistency}

To assess deployment fidelity, we evaluate whether MobileAgeNet preserves its age-regression behavior after export from ONNX to TensorFlow Lite (TFLite). The objective of this experiment is conversion consistency rather than post-export performance improvement. Because our framework targets mobile and edge deployment, this analysis provides a practical validation of model portability.

The consistency analysis follows the same age-stratified split policy used in training (seed 42). ONNX and TFLite models are evaluated on the same ordered sample set for each split, using identical post-export preprocessing: RGB conversion, resize to $224 \times 224$, and ImageNet normalization with mean $(0.485, 0.456, 0.406)$ and standard deviation $(0.229, 0.224, 0.225)$. ONNX inference is executed with ONNX Runtime on CPU, while TFLite inference uses the TensorFlow Lite interpreter with the XNNPACK CPU backend. Dynamic input dimensions are resized explicitly before inference when required.

Table~\ref{tab:onnx_tflite_consistency} reports results on the full validation split ($N=2372$). ONNX and TFLite MAE are numerically identical at 4.8387 years, indicating no measurable degradation introduced by conversion under the evaluation protocol.

To quantify conversion fidelity, we compute:
\begin{equation}
\Delta_{\mathrm{conv}} = \left| \mathrm{MAE}_{\mathrm{ONNX}} - \mathrm{MAE}_{\mathrm{TFLite}} \right|.
\end{equation}
In our case, $\Delta_{\mathrm{conv}} = 0.0000$, which indicates perfect agreement between ONNX and TFLite at the MAE level on the same validation samples.

We also compare post-conversion validation performance to the best validation MAE recorded during training:
\begin{equation}
\Delta_{\mathrm{val}} = \left| \mathrm{MAE}_{\mathrm{TFLite,val}} - \mathrm{MAE}_{\mathrm{best\_val,train}} \right|.
\end{equation}
We obtain $\Delta_{\mathrm{val}} = 0.0864$ years. This small discrepancy is expected and reflects differences between training-time validation and post-export runtime evaluation. In the current pipeline, training-time validation reuses the selected training transform configuration (which includes stochastic augmentations), whereas post-export ONNX/TFLite evaluation uses deterministic resize-and-normalize preprocessing.

Importantly, cross-split comparisons are not used to assess conversion fidelity. Held-out results are reported separately from validation results to avoid interpretation errors. Therefore, the deployment conclusion is based on same-split ONNX--TFLite parity, which demonstrates stable behavior of the exported TFLite model for downstream mobile benchmarking and deployment-oriented evaluation.

\begin{table}[t]
\centering
\small
\caption{ONNX-to-TFLite consistency on the UTKFace validation split (age-stratified, seed 42). Lower is better for MAE.}
\label{tab:onnx_tflite_consistency}
\begin{tabular}{lcc}
\toprule
Metric & Value & Unit \\
\midrule
Best validation MAE during training & 4.7523 & years \\
ONNX MAE (validation evaluation) & 4.8387 & years \\
TFLite MAE (post-conversion validation) & 4.8387 & years \\
$\Delta_{\mathrm{val}}$ (TFLite vs best training validation) & 0.0864 & years \\
Evaluated samples & 2372 & images \\
TFLite model size & 12.34 & MB \\
\bottomrule
\end{tabular}
\end{table}

\subsection{On-Device Latency Analysis}
\label{subsec:mobile_latency}

To evaluate practical deployment efficiency, we measured on-device inference latency using the AI Benchmark application on a real mobile device (Motorola Edge 40). Unlike low-level benchmarking tools that report kernel execution time, AI Benchmark provides a more realistic estimate of inference latency by incorporating runtime overheads such as tensor preparation, scheduling, and memory access.

The model was evaluated with an input resolution of $224 \times 224 \times 3$, and latency statistics were collected over 20 consecutive inference runs. Table~\ref{tab:mobile_latency} summarizes the initialization time, average latency, and standard deviation.

The model achieves an average on-device inference latency of \textbf{14.4 ms}, with a low standard deviation of \textbf{1.4 ms}, indicating stable runtime performance. The initialization time is \textbf{25.6 ms}, which is incurred only once during model loading. These results correspond to practical end-to-end inference latency under realistic mobile deployment conditions.

For completeness, we also evaluated the model using Google's LiteRT benchmark tool, which reports steady-state kernel execution latency under optimized conditions. While the LiteRT benchmark indicated sub-millisecond inference latency, such measurements exclude preprocessing, memory overhead, and runtime scheduling, and therefore do not reflect full application-level latency. We therefore report AI Benchmark results as the primary deployment metric.

In summary, the proposed MobileAgeNet achieves real-time performance on a consumer smartphone, with inference latency well below the typical interactive threshold of 30 ms, making it suitable for mobile age estimation applications.

\begin{table}[t]
\centering
\small
\caption{On-device inference latency of the proposed model on a Motorola Edge 40 measured using AI Benchmark. Lower is better.}
\label{tab:mobile_latency}
\setlength{\tabcolsep}{6pt}
\begin{tabular}{lccc}
\toprule
Metric & Value (ms) \\
\midrule
Initialization Time & 25.6 \\
Average Latency & \textbf{14.4} \\
Latency Std. Dev. & 1.4 \\
\bottomrule
\end{tabular}
\end{table}

\begin{figure}[t]
    \centering
    \includegraphics[width=0.9\columnwidth]{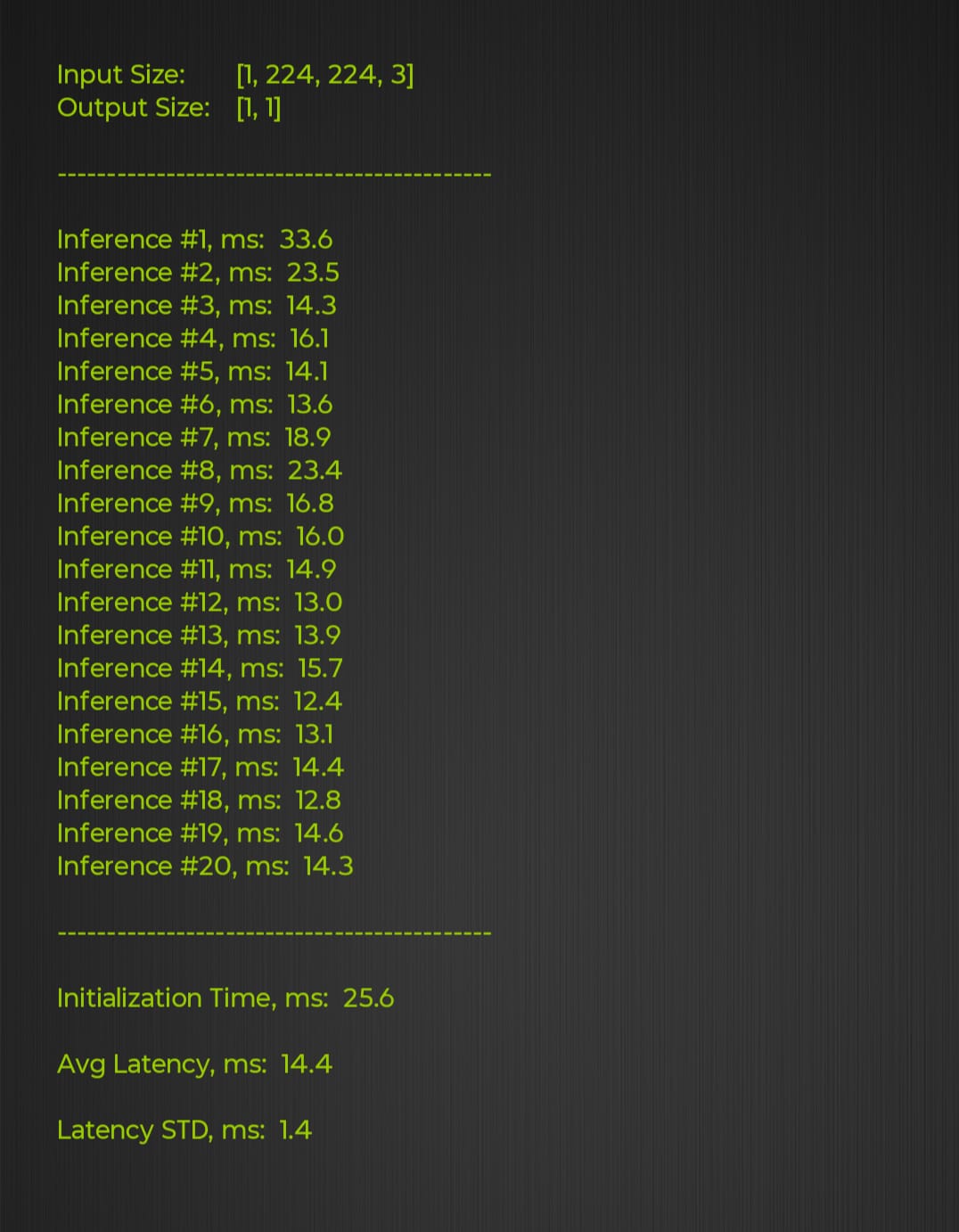}
    \caption{Latency measurement results obtained using the AI Benchmark application on the Motorola Edge 40. The model processes $224 \times 224$ input images and achieves stable inference latency across multiple runs.}
    \label{fig:latency_test}
    \vspace{-2mm}
\end{figure}
\section{Conclusion}
\label{sec:conclusion}

MobileAgeNet demonstrates that a pretrained MobileNetV3-Large backbone 
combined with a compact regression head, bounded age supervision, and a 
two-stage fine-tuning strategy is sufficient to achieve competitive facial 
age estimation under a fully reproducible evaluation protocol. The proposed 
model attains a held-out MAE of 4.65 years on the full-range UTKFace test 
set at only 3.23M parameters and 0.233~GFLOPs, outperforming prior 
MobileNet-based baselines on the same dataset~\cite{Savchenko2021Lightweight} and 
matching distilled MobileNetV3-Large results obtained on a substantially 
easier restricted age range~\cite{Kim2024KD}.

On-device evaluation on a Motorola Edge~40 via AI~Benchmark confirms 
stable inference at 14.4~ms average latency with a standard deviation of 
only 1.4~ms, well within the 30~ms interactive threshold. Conversion 
consistency analysis further shows zero measurable MAE degradation 
across the PyTorch$\,{\to}\,$ONNX
$\,{\to}\,$TFLite pipeline, confirming deployment fidelity without 
post-export accuracy loss.

The full pipeline is integrated into the NN LEMUR Dataset 
framework~\cite{Goodarzi2025LEMUR}, providing structured hyperparameter 
optimization via Optuna, validation-based checkpoint selection, and 
age-stratified held-out evaluation --- a reproducibility standard that 
remains uncommon in age-estimation literature~\cite{Paplham2024Benchmark}. 
Together, these results position MobileAgeNet as a transparent, 
deployment-ready baseline for mobile facial age estimation. Future work 
will explore knowledge distillation, INT8 quantization-aware training, 
and extension to additional facial analysis tasks within the same 
framework.

\vspace{0.2cm}
\noindent\textbf{Acknowledgments.}
This work was partially supported by the Alexander von Humboldt Foundation.

	{
		\small
		\bibliographystyle{ieeenat_fullname}
		\bibliography{bibmain}
	}
	
\end{document}